%% file: T-Prompter.tex
\documentclass[acmtog,nonacm]{acmart}
\acmSubmissionID{599}

\usepackage{booktabs} %
\usepackage{microtype}
\usepackage{xspace}
\usepackage{enumitem}

\input{macros}

\usepackage[ruled]{algorithm2e} %

\SetAlFnt{\small}
\SetAlCapFnt{\small}
\SetAlCapNameFnt{\small}
\SetAlCapHSkip{0pt}
\acmJournal{TOG}

\newcommand{\hide}[1]{}

\begin{document}
\title{IP-Prompter: Training-Free Theme-Specific Image Generation via Dynamic Visual Prompting}

\author{Yuxin Zhang}
\orcid{0000-0001-6433-2678}
\author{Minyan Luo}
\orcid{0009-0008-2675-1650}
\author{Weiming Dong}
\authornote{Corresponding author: Weiming Dong (weiming.dong@ia.ac.cn)}
\orcid{0000-0001-6502-145X}
\affiliation{
\institution{MAIS, Institute of Automation, CAS}
\country{China}
}
\affiliation{
\institution{School of Artificial Intelligence, UCAS}
\country{China}
}
\email{zhangyuxin2020@ia.ac.cn}
\email{luominyan21@mails.ucas.ac.cn}
\email{weiming.dong@ia.ac.cn}

\author{Xiao Yang}
\orcid{0009-0007-2411-3594}
\author{\, Haibin Huang}
\orcid{0000-0002-7787-6428}
\author{Chongyang Ma}
\orcid{0000-0002-8243-9513}
\affiliation{
\institution{ByteDance Inc.}
\country{China}
}
\email{yangxiao.0@bytedance.com}
\email{jackiehuanghaibin@gmail.com}
\email{chongyangm@gmail.com}

\author{Oliver Deussen}
\orcid{0000-0001-5803-2185}
\affiliation{
\institution{University of Konstanz}
\country{Germany}
}
\email{oliver.deussen@uni-konstanz.de}

\author{Tong-Yee Lee}
\orcid{0000-0001-6699-2944}
\affiliation{
\institution{National Cheng-Kung University}
\country{Taiwan}
}
\email{tonylee@ncku.edu.tw}

\author{Changsheng Xu}
\orcid{0000-0001-8343-9665}
\affiliation{
\institution{MAIS, Institute of Automation, CAS}
\country{China}
}
\affiliation{
\institution{School of Artificial Intelligence, UCAS}
\country{China}
}
\email{csxu@nlpr.ia.ac.cn}

\renewcommand\shortauthors{Zhang et al.}

\input{Sections/0-Abstract}



\begin{teaserfigure}
    \centering
    \includegraphics[width=\linewidth]{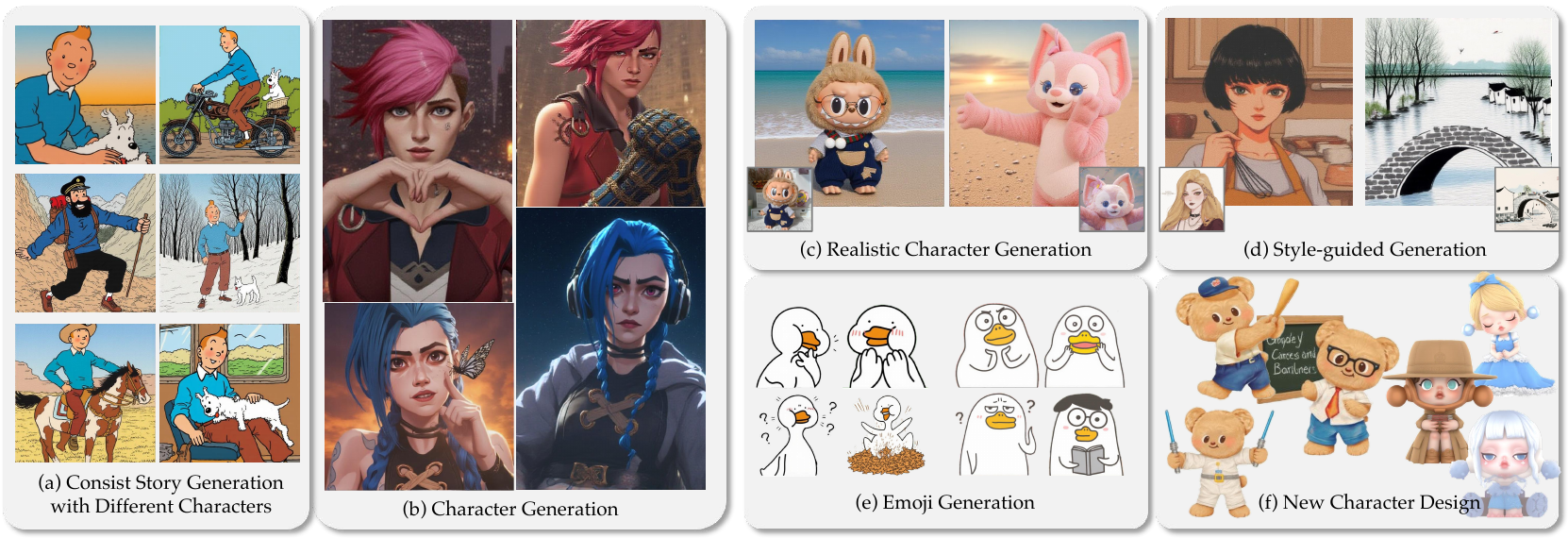}
    \caption{Training-free generation results of \sysname.
    Each result presented in this paper is generated using a fixed random seed.}
    \label{fig:teaser}
\end{teaserfigure}
\maketitle

\input{Sections/1-Introduction}

\input{Sections/2-Related}

\input{Sections/3-Method}

\input{Sections/4-Experiment}

\input{Sections/5-Application}

\input{Sections/6-Conclusion}

\input{Figs/fig_supp}
\begin{acks}
This work was supported in part by the Beijing Natural
Science Foundation under nos. L221013 and QY24384, in part by the Beijing Training Program of Innovation and Entrepreneurship for Undergraduates under no. S202414430024, in part by the National Science and Technology Council under no. 111-2221-E-006-112-MY3, Taiwan, in part by the Deutsche Forschungsgemeinschaft (DFG, German Research Foundation) under Germany's Excellence Strategy-EXC 2117-422037984.
\end{acks}

\bibliographystyle{ACM-Reference-Format}
\bibliography{T-Prompter}

\end{document}

%% file: macros.tex
\usepackage{ifthen}
\usepackage{soul}

\newcommand{\sysname}{\textit{IP-Prompter}\xspace}
\newcommand{\sysnames}{\textit{IP-Prompter's}\xspace}

\usepackage{ifthen}
\definecolor{WeimingColor}{rgb}{0,0,0.8} 
\definecolor{OliverColor}{rgb}{0.8,0,0.8}
\newcommand{\weiming}[1]{{\color{WeimingColor} [Weiming: #1]}}

\newcommand{\revision}[1]{#1}

\newcommand{\final}{1}

\ifthenelse{\equal{\final}{1}}
{
\renewcommand{\weiming}[1]{}

}

%% file: Sections/0-Abstract.tex
\begin{abstract}
    The stories and characters that captivate us %
    as we grow up shape unique fantasy worlds, 
    with images serving as %
    the primary medium for visually experiencing these realms. %
    Personalizing generative models through fine-tuning with theme-specific data has become a prevalent approach in text-to-image generation.
    However, unlike object customization, which focuses on learning specific objects, theme-specific generation encompasses diverse elements such as characters, scenes, and objects. Such diversity also introduces a key challenge: how to adaptively generate multi-character, multi-concept, and continuous theme-specific images (TSI). Moreover, fine-tuning approaches often come with significant computational overhead, time costs, and risks of overfitting.
    This paper explores a fundamental question: Can image generation models directly leverage images as contextual input, similarly to how large language models use text as context? 
    To address this, we present \sysname, a novel training-free TSI generation method. 
    \sysname introduces visual prompting, a mechanism that integrates reference images into generative models, allowing users to seamlessly specify the target theme without requiring additional training. To further enhance this process, we propose a Dynamic Visual Prompting (DVP) mechanism, which iteratively optimizes visual prompts to improve the accuracy and quality of generated images.
    Our approach enables diverse applications, including consistent story generation, character design, realistic character generation, and style-guided image generation. 
    Comparative evaluations against state-of-the-art personalization methods demonstrate that \sysname achieves significantly better results and excels in maintaining character identity preserving, style consistency and text alignment, offering a robust and flexible solution for theme-specific image generation.
    Our project page: \url{https://ip-prompter.github.io/}.
\end{abstract}

%% file: Sections/1-Introduction.tex
\section{Introduction}

If a picture is worth a thousand words, then a theme tells an entire story. We define a theme-specific image (TSI) as a visual composition that cohesively integrates characters, objects, and environments within a unified artistic style or narrative framework. TSIs align explicitly with a defined theme or concept, making them essential for thematic communication and audience engagement. These images have applications in areas such as branding, storytelling, and design.
Recent advancements in text-to-image generation models have enabled users to synthesize images from text prompts. For generating images of specific concepts, popular methods include personalized techniques such as model fine-tuning~\cite{ruiz2023dreambooth}, the integration of auxiliary control networks~\cite{mou2024t2i,ControlNet}, and attention exchange mechanisms for concept injection~\cite{chung2024style,hertz2024style}. However, unlike object customization tasks, which focus on learning specific objects, generating TSIs involves managing a diverse set of elements, including characters, scenes, and objects. For instance, as shown in Fig.~\ref{fig:teaser}(a), TSIs of \textit{The Adventures of Tintin} encompass multiple elements such as Tintin, Snowy, and Captain Haddock. This diversity presents a significant challenge: designing methods capable of flexibly and efficiently adapting to multi-character, multi-concept, and continuous generation tasks, all while minimizing costs. Existing fine-tuning approaches struggle with rapid concept switching and introduce high computational and time overhead. Similarly, approaches that rely on auxiliary networks or attention exchange mechanisms often struggle to maintain the identity consistency of characters and objects while necessitating structural modifications to large pre-trained models, thereby introducing additional challenges.

On the other hand, large language models (LLMs) have demonstrated the ability to use user-provided context as knowledge, offering convenience in textual communication. Inspired by this, we extend this paradigm to visual communication in image generation models. Despite the inherent challenges of TSI generation, it offers a unique advantage: the availability of extensive visual context in the form of character, object, and background images, which can be leveraged directly. In this paper, we introduce a visual prompting interaction framework based on image inpainting models. As illustrated in Fig.~\ref{fig:motivation}(c), this framework incorporates personalized concepts by directly stitching guiding images as contextual information. Visual prompting eliminates the need for additional networks, modules, or attention mechanisms. This training-free and modification-free approach not only enhances efficiency but also ensures that guidance information remains within the same visual domain as the target output, resulting in precise results.

To address the core challenges of TSI generation, we further propose a Dynamic Visual Prompting (DVP) scheme.
DVP matches and arranges visual prompts in real time based on the target theme and user-provided text instructions.
Users only need to provide a thematic image collection, while DVP enables precise control over the generation model.
The key steps of DVP include analyzing user intentions, matching context information, composing visual prompts, iterative updating, and evaluation. 
First, DVP automatically extracts key visual elements from the user’s input text prompts. Subsequently, it performs visual-textual matching within the theme-specific image collection.
The images with the highest matching scores are then composed into visual prompts in a specific arrangement according to an importance assessment.
These visual prompts are then updated in self-consistency manner and fed into the generation model.
In the last stage, the generated results are evaluated and satisfactory output is returned to the user.
By dynamically adjusting reference images according to user instructions, DVP ensures superior flexibility, accuracy, and efficiency. We name our TSI generation method incorporating DVP as \sysname.

\input{Figs/fig_motivation}

We also conduct extensive comparisons with a wide range of baseline methods on TSI generation tasks. \sysname demonstrates state-of-the-arts performance in theme consistency and text-image alignment. Moreover, \sysname can directly assist users in diverse design applications, including consistent story generation, character design, realistic character generation, and style-guided image creation.
In summary, our contributions are as follows.
\begin{itemize}[leftmargin=*,topsep=0pt]

\item  \revision{We introduce a new approach for visual prompting, i.e., interacting with the image generation model in a training-free and modification-free manner through a multi-grid format.
We emphasize the importance of optimizing visual prompts in the same way as carefully designing text prompts.
}

\item  We propose \sysname, a TSI generation method that leverages DVP to dynamically adjust guiding images according to user input. DVP tackles the core challenge of flexibility in TSI generation, and offers exceptional accuracy and efficiency, seamlessly adapting to multi-turn dialogues.

\item Extensive experiments demonstrate that \sysname achieves state-of-the-art performance in TSI generation tasks. Additionally, \sysname supports diverse creative applications, including consistent story generation, character design, realistic character generation, and style-guided image creation.

\end{itemize}

%% file: Figs/fig_motivation.tex
\begin{figure*}
\centering
\includegraphics[width=1\linewidth]{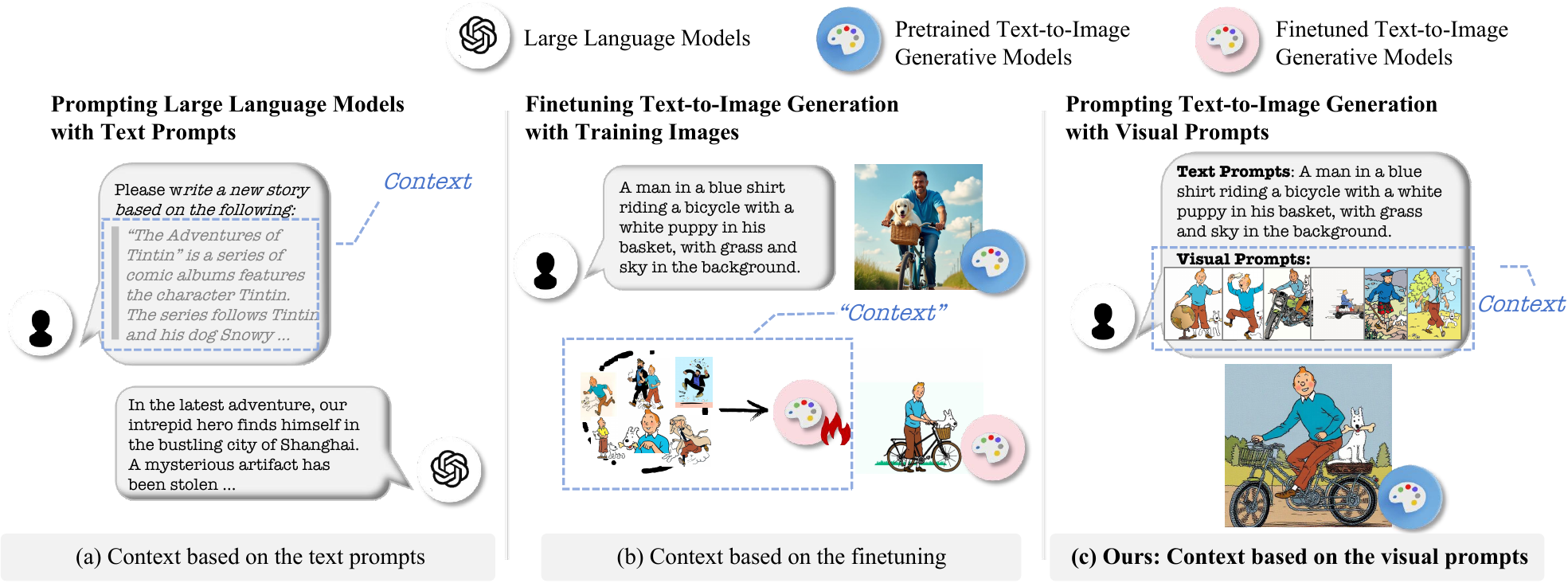}
\caption{\textbf{Schematic illustration of visual promoting:} (a) Text prompting in LLMs provides context and knowledge for the model to generate target content. (b) Existing personalized methods inject concepts into the model by fine-tuning the model, training a reference network, or altering the model structure to achieve thematic control. (c) Our proposed visual prompting based on inpainting represents a new model interaction paradigm, where visual prompts directly provides contextual information to the model, enabling fast and efficient controllable generation without the need to modify the generative model.
}
\label{fig:motivation}
\end{figure*}

%% file: Sections/2-Related.tex
\section{Related Work}

Theme-specific and consistency-oriented image generation tasks focus on creating a series of new images that preserve consistent visual and semantic characteristics guided by input directives. Achieving this goal necessitates models capable of reliably maintaining the core identity of personalized content while adapting it to varied contexts. Recent progress in this area has led to the development of diverse approaches, broadly classified into training-based methods and training-free image guidance techniques.

Conventional approaches to personalized generation often depend on fine-tuning models or incorporating supplementary networks to attain precise control over the generated outputs.
\revision{
Techniques such as DreamBooth~\cite{ruiz2023dreambooth}, LoRA~\cite{hu2021lora}, and subsequent works~\cite{customdiff,chen2024anydoor,shah2025ziplora,sohn2024styledrop,xu2024break,purushwalkam2024bootpig,jang2024identity} focus on fine-tuning the model’s attention mechanisms to embed personalized concepts effectively.}
\revision{
Another class of methods, including Textual Inversion~\cite{gal2023TI} and subsequent works~\cite{wei2023elite,zhang2023inversion,zhang2023prospect,avrahami2023break,vinker2023concept,huang2024reversion,li2024photomaker,zeng2024infusion}, invert guiding images into the textual space, enabling concept guidance through textual prompts.  }
\revision{
Frameworks like T2I-Adapter~\cite{mou2024t2i}, ControlNet~\cite{ControlNet}, and related approaches~\cite{ip-adapter,parmar2025object,wang2024instantstyle,wang2024instantid,easyref,gal2024lcm} incorporate additional reference networks to encode and inject guidance concepts directly into the generation pipeline.}
In particular, these methods do not require additional fine-tuning during inference.
While these approaches facilitate personalized generation, they typically require extensive parameter updates and incur substantial computational overhead, limiting their efficiency in scenarios that demand scalability or rapid adaptation.

Recent trends in personalized generation focus on training-free methods, which achieve concept injection by manipulating features or utilizing the capabilities of pre-trained models.
For instance, StyleAligned~\cite{hertz2024style} and concurrent techniques~\cite{chung2024style, Deng:2024:ZST,alaluf2024cross} achieve style-consistent image generation by swapping keys and values between reference and generated images in the attention layers.
ConsiStory~\cite{consistory} introduces a subject-driven shared attention block and correspondence-based feature injection mechanism.
FreeCustom~\cite{freecustom} introduces a multi-reference self-attention  mechanism and a weighted mask strategy to inject concepts.
The above methods requires manipulations of model structures and are challenging generation characters of different actions.

\revision{
Some methods explore visual prompting in the field of computer vision. Bar et al.~\shortcite{bar2022visual} propose that various computer vision tasks can be treated as grid inpainting problems. They also highlight the importance of diverse data.
Zhang et al.~\shortcite{zhang2023makes} conduct an investigation on the impact of in-context examples in computer vision and found that the performance is highly sensitive to the choice of examples.
In the field of image generation, some methods~\cite{yeh2024gen4gen} utilize image prompting to achieve guidance. Analogist~\cite{gu2024analogist} explores the performance of visual in-context learning by proposing self-attention cloning.
IC-LoRA~\cite{huang2024context} maps reference images to generate output via LoRA training. Group diffusion transformers~\cite{huang2024group} establishes associations using a group-attention block. JeDI~\cite{zeng2024jedi} injects reference information through coupled self-attention. OminiControl~\cite{tan2024ominicontrol} proposes a minimal control framework to achieve conditioning. Diptych Prompting~\cite{shin2024large} leverages the inherent consistency capabilities of pre-trained models and integrates a single reference image via attention re-weighting.
In contrast, we introduce prompt engineering to the visual domain by treating the dataset as an independent prompt bank that provides visual elements, rather than embedding dataset-specific information into model parameters.
}

%% file: Sections/3-Method.tex
\section{Method}
\input{Figs/fig_pipeline}

Dynamic Visual Prompting (DVP) is designed  to generate novel images that align with a user-provided image set while adhering to the content of a textual prompt. Inspired by human creative processes, theme-specific creation typically involves defining the subject, designing its appearance, iteratively refining the design, and producing high-quality outputs.
Therefore, the challenges in theme-specific generation can be distilled into the following key tasks:
(1) Clarifying the subject of creation: How can user intentions be effectively decomposed and interpreted to produce meaningful outputs?
(2) Establishing textual-visual connections: How can textual prompts be linked to visual elements to ensure generated outputs reflect the desired content and appearance?
(3) Optimizing generation results: How can visual prompts be organized and selected to maximize output quality? Furthermore, how can the generated results be systematically evaluated?

To address these challenges, we propose DVP (see Fig.\ref{fig:pipeline}), a framework composed of three steps: (1) user intent understanding and key element extraction; (2) visual prompt matching and generation; and (3) self-consistent prompt updating and evaluation. 
DVP is designed to (4) seamlessly switch between different creative subjects, providing enhanced flexibility and efficiency in content generation.

\paragraph{User Intent Understanding and Key Element Extraction}

The first step in DVP is to interpret the user's creative intentions and identify the central subject of the creation.
To accomplish this, DVP employs visual element extraction based on the user’s input textual prompt.
This process can be performed either automatically, using LLMs, or manually, through user input. Specifically, we utilize a structured text command structured as follows:
``\textit{Please extract $N$ key visual elements from this paragraph.}''
Here, $N$ is either set to a default value (e.g., $N=3$) or specified by the user. This process yields a set of key elements \( \{element_0, \dots, element_N\} \).
Identifying these elements is critical, as failing to do so may result in visual prompts lacking the specific semantic details necessary for accurate and meaningful content generation.

\paragraph{Visual Prompt Matching and Generation}

\input{Figs/fig_attention}
\input{Figs/fig_comparison}

The second step establishes a connection between the textual and visual prompts by leveraging the CLIP model~\cite{clip} to map those elements into a shared embedding space.
The CLIP text encoder and image encoder are used to transform the extracted key elements and images into text embeddings \( \{E_0, \dots, E_N\} \), and image embeddings \( \{I_0, \dots, I_M\} \), where $M$ corresponds to the number of user-provided images.
The similarity between each text embedding and image embedding is then computed as:
$similarity_{i,j} = \frac{E_i \cdot I_j}{\|E_i\| \|I_j\|}$.
For each visual element, the top-$K$ images with the highest similarity scores are selected, resulting in \(N \times K\) image candidates. 
We set \(N=3\) and \(K=3\), and this process yields a $3\times3$ image template (see Fig.~\ref{fig:pipeline}).

The arrangement of visual prompts within the mask-filling generative model plays a crucial role in the generation process.
To analyze this influence, we conduct attention visualization experiments.
For the same group of reference images, inference is performed with different grid arrangement.
We calculate the average attention maps (annular) of the reference image to the generated image obtained during all inferences, and the results are shown in the upper right corner of Fig.~\ref{fig:attention}.
This reveals that specific starred regions exhibit higher average attention intensities (deeper colors) compared to other areas, indicating their significance within the image grid.
Consequently, the most important images are placed in these high-attention regions, and are complemented by a central mask. These inputs are then fed into an image inpainting model, which synthesizes the target image within the central mask, guided by the user-provided textual input.
\revision{We use six arrangements in all experiments. Three visual elements are fully permuted across rows.}

\paragraph{Self-consistent Prompt Updating and Evaluation}

For images within the starred/unstarred region, variations in images arrangement can still influence the generated images. 
Similar to the behavior observed in LLMs, where semantically identical prompts may yield divergent outputs, a self-consistent strategy~\cite{wang2022self} was adopted to address this variability. 
In LLMs, self-consistency typically involves evaluating outputs generated from different prompt arrangements and selecting the most plausible result through a voting mechanism.
Inspired by dynamic prompting~\cite{yang2023dynamic} and self-consistency in LLMs, we introduced a self-consistent prompt updating approach.
As shown in the iterative refinement stage of Fig.~\ref{fig:pipeline}, the model iteratively rearranges a set of visual prompts to produce diverse outputs.
The best match to the user’s requirements is then selected using quantitative metrics, including text-image consistency (CLIP text score), thematic consistency (CLIP image score), and visual quality (evaluated using visual-language models).
In this way, DVP can help users obtain the most suitable combination of reference images and reduce their burden on selection.

%% file: Figs/fig_pipeline.tex
\begin{figure*}
\centering
\includegraphics[width=1\linewidth]{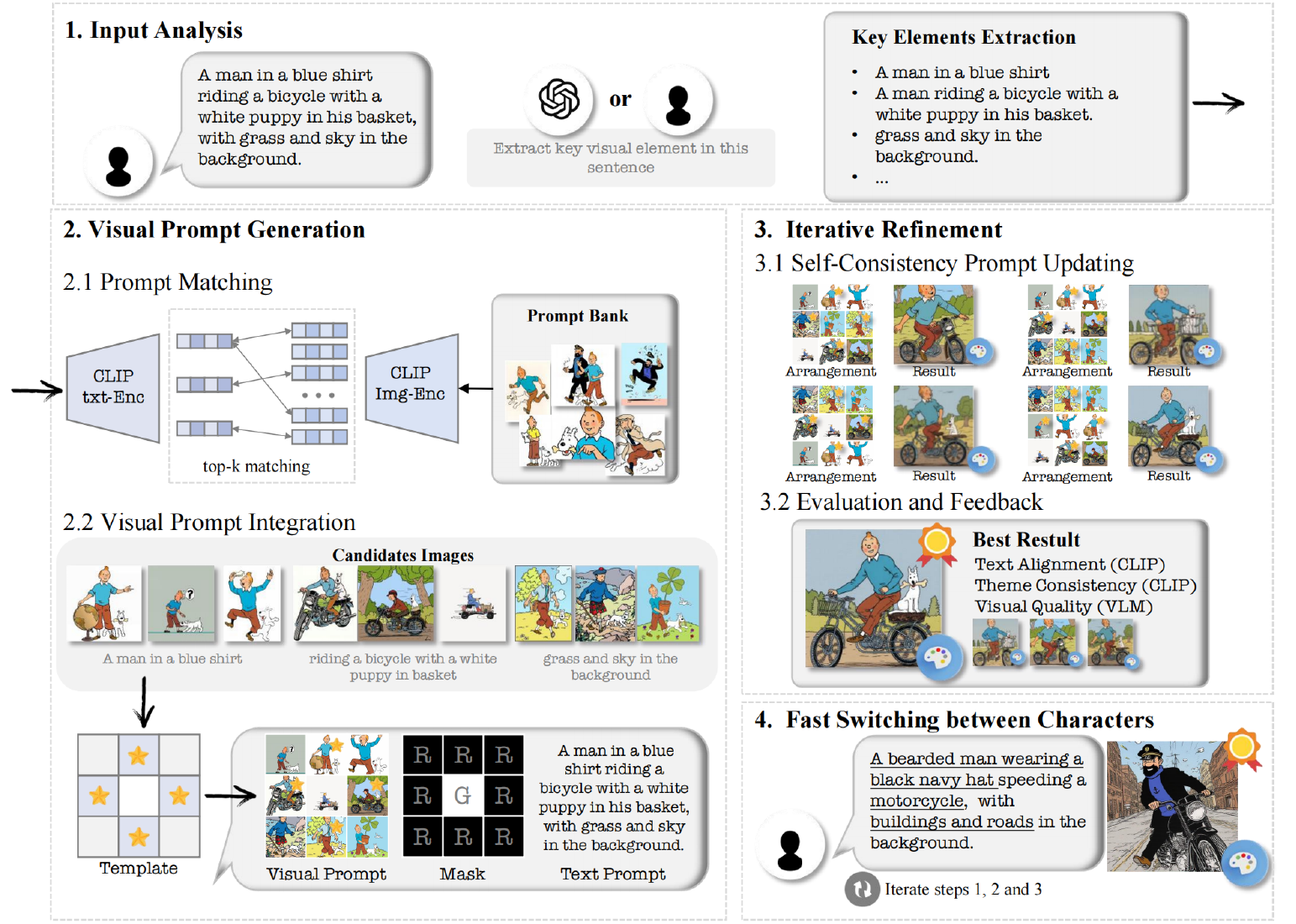}
\caption{\textbf{Pipeline of \sysname:}
Dynamic visual prompting (DVP) includes three key stages: (1) Comprehending user intent and extracting key elements; (2) Matching and generating visual prompts; and (3) Updating and evaluating prompts through self-consistency. 
This way (4) DVP enables effortless transition between diverse creative subjects, thereby enhancing the flexibility and efficiency of content generation.}
\label{fig:pipeline}
\end{figure*}

%% file: Figs/fig_attention.tex
\begin{figure}
\centering
\includegraphics[width=1\linewidth]{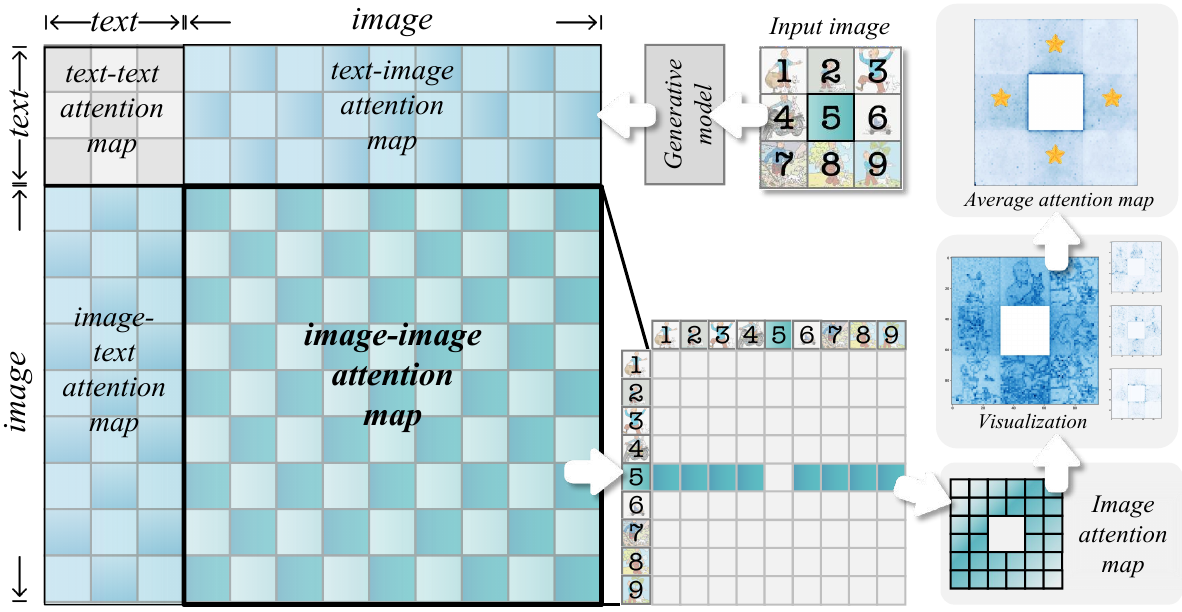}
\caption{%
\textbf{Attention maps computed during model inference}. Here, a lots of visual prompts are combined in various arrangements. As shown in the upper right corner, areas with deeper colors are allocated more attention. These are referred to as significant regions and are marked with star symbols.
}
\label{fig:attention}
\end{figure}

%% file: Figs/fig_comparison.tex
\begin{figure*}
\centering
\includegraphics[width=\linewidth]{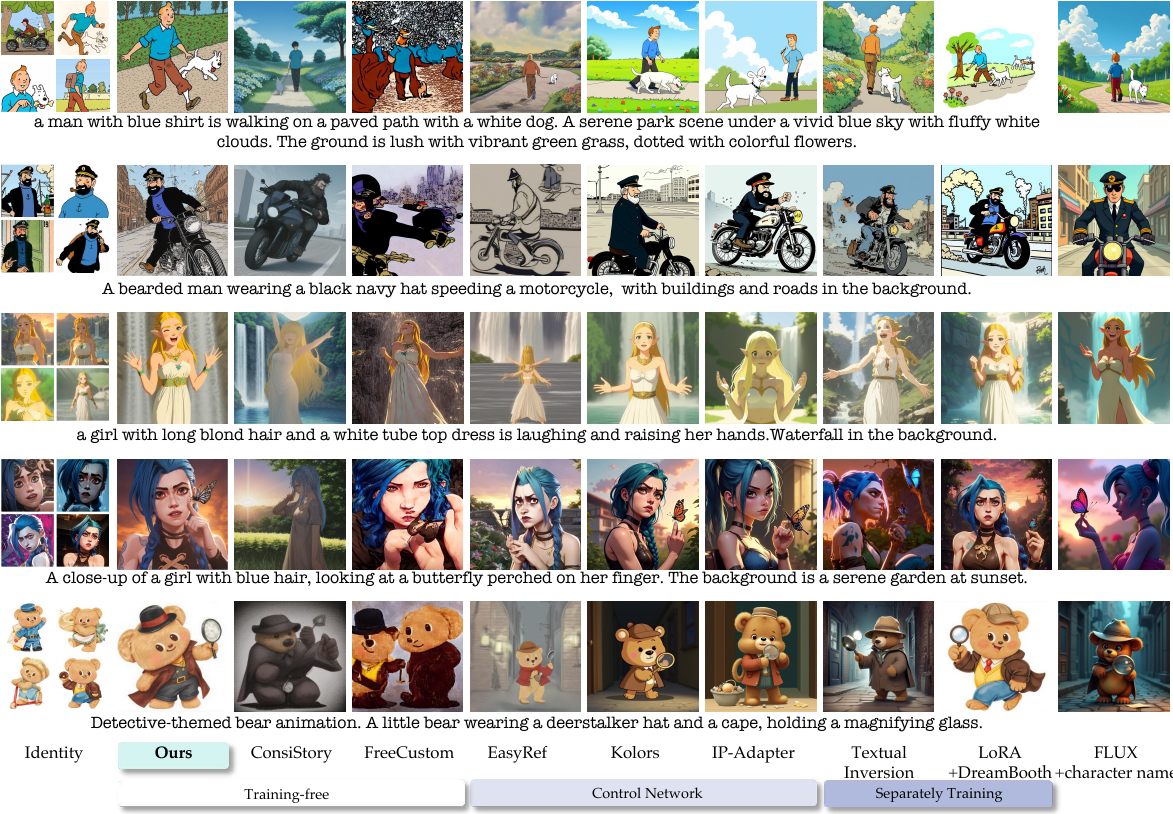}
\caption{\textbf{Qualitative Results.} We compare \sysname with the SOTA personalization methods, including FLUX 1.0, Textual Inversion (TI) with SDXL, DreamBooth+LoRA with FLUX, Kolors Character with FLUX, IP-Adapter with FLUX, EasyRef, FreeCustom and ConsiStory.
}
\label{fig:comparison}
\end{figure*}

%% file: Sections/4-Experiment.tex
\section{Experiments}

\paragraph{Evaluation baselines}

We compared our method with state-of-the-arts personalization methods, including FLUX 1.0~\cite{flux}, Textual Inversion (TI)~\cite{gal2023TI} with SDXL~\cite{sdxl}, DreamBooth+LoRA~\cite{ruiz2023dreambooth} with FLUX, Kolors~\cite{kolors} Character with FLUX, IP-Adapter~\cite{ip-adapter} with FLUX, EasyRef~\cite{easyref}, FreeCustom~\cite{freecustom}, and ConsiStory~\cite{consistory}.

\paragraph{Implementation details}
We used FLUX-Fill 1.0~\cite{flux} with the default hyper-parameters in all our experiments. 
The guidance scale is 30 and the number of inference steps is 50.
\revision{
\sysname generates strong results with 15 diverse images per character and excels with 30.
The matching and evaluation process take about 10 seconds.
The synthesis process takes about 30 seconds for a $512\times 512$ image, which is comparable with baseline training-free methods and faster than minute-level finetuning-based methods and day-level adapter-based methods.
Training-based methods like LoRA~\cite{hu2021lora} offer higher adaptability and precision at the cost of additional setup, whereas training-free methods prioritize convenience and general applicability.}

\subsection{Qualitative Evaluation}
We compare our method with eight state-of-the-arts personalization methods.
As illustrated in Fig.~\ref{fig:comparison}, we categorize the evaluated methods into three groups: training-free methods, methods that require training an additional control network, and methods that involve separate learning for each concept.
Notably, except for FLUX, none of the other methods explicitly leverages thematic information.

ConsiStory~\cite{consistory} demonstrates the capability to generate a series of consistent images in text-to-image scenarios but struggles to achieve precise control in image-guided generation tasks. FreeCustom~\cite{freecustom} is able to generate consistent objects, but has difficulty with actions and background changes
EasyRef~\cite{easyref} utilizes the representation capabilities of multi-modal models and produces favorable results in certain scenarios, as shown in the $4^\text{th}$ row of Fig.~\ref{fig:comparison}. However, it fails to preserve identity information when dealing with non-human domains, such as cartoons and animals.
Kolors-Character~\cite{kolors} and IP-Adapter~\cite{ip-adapter} are based on single-image guidance, effectively captures basic image characteristics, such as cartoon styles, and exhibits strong text-image consistency. However, they struggle with precise character control.
For example, Kolors fails to preserve character identity in the $1^\text{st}$, $2^\text{nd}$, $5^\text{th}$, and $6^\text{th}$ rows, while IP-Adapter fails in the $1^\text{st}$, $3^\text{rd}$, $5^\text{th}$, and $6^\text{th}$ rows.
TI ~\cite{gal2023TI} achieves commendable generation quality and text-image consistency, but struggles to capture intricate character appearances, particularly when character actions vary significantly.
DreamBooth+LoRA~\cite{ruiz2023dreambooth} delivers the best results among the baseline methods. However, slight compromises in overall style and character identity are noticeable in the $1^\text{st}$ and $2^\text{nd}$ rows of Fig.~\ref{fig:comparison}.
In contrast, as demonstrated in the $2^\text{nd}$ column of Fig.~\ref{fig:comparison}, our method achieves superior thematic and text-image consistency. \sysname effectively controls diverse character types, enabling them to perform dynamic actions and appear in novel scenes, while preserving fine-grained details, such as clothing logos.
Notably, the results of \sysname are difficult to distinguish from the real ones.
For fair comparison, all results are generated using the a single random seed.

\subsection{Quantitative Evaluation}

\input{Figs/fig_quantitative}
\input{Figs/fig_ablation}
\input{Figs/fig_app}
\input{Figs/fig_story}
\input{Figs/fig_story2}

We employ two metrics for quantitative evaluation.
We select ten familiar themes.
For each theme, we use five prompts, generating ten images per prompt with different random seeds, resulting in a total of 500 images for each method.
As shown in Fig.~\ref{fig:quantitative}, the vertical axis represents the \emph{image similarity}, measured as the pairwise CLIP cosine similarity between the reference images and the generated images. The higher the better thematic fidelity.
The horizon axis represents the \emph{text similarity}, measured as the CLIP similarity between all generated images and their textual conditions. The higher the better editability.
Our method achieves the highest thematic consistency while maintaining a text instruction-following capability comparable to FLUX. 
\sysname achieves performance comparable to the fine-tuned FLUX model using the DreamBooth+LoRA approach.
For user study, we invited 50 participants to rate the results generated by each method for 10 common themes on a scale of 0 to 5, and the results are shown in the right part of Fig.~\ref{fig:quantitative}. \sysname received the highest user preference.

\subsection{Ablation Study}

\revision{As shown in Figs.~\ref{fig:quantitative} and \ref{fig:ablation}(a), we conduct an ablation study on different steps in DVP.
The ablation of each component contain 100 generated results.}
Without the self-consistency prompt updating process, it is difficult for the model to match the most suitable image in the most important position, resulting in a loss of detail, such as the absence of the red coat in the $2^\text{nd}$ row. 
Without attention-based arrangement, this detail loss is exacerbated, for instance, the identity in the $2^\text{nd}$ row loses. 
Without prompt matching, the guiding image and the target content are misaligned, which causes inconsistencies in the characters. 
Finally, without any DVP mechanism, it is difficult to generate images with consistent style and characters, demonstrating the capability of the visual prompting mechanism.

As shown in Fig.~\ref{fig:ablation}(b), we conducted an ablation study on different numbers and formats of visual prompting. The black regions in the mask indicate the position of the reference image, while the white regions represent the canvas generated by the model. From left to right, the configurations correspond to nine, four, and two reference images, respectively. It is evident that as the number of reference images decreases, the generated results suffer from style degradation and even character identity loss. 
A reduced number of reference images fails to provide sufficient context for the model, and this lack of diversity causes the model to ignore the influence of the reference images when generating new content.
\revision{
The quantitative ablation study results in Fig.~\ref{fig:quantitative} demonstrate the importance of the number of reference images in maintaining character consistency. Reducing the grid size results in a significant decrease in image scores.}

%% file: Figs/fig_quantitative.tex
\begin{figure}
\centering
\includegraphics[width=1\linewidth]{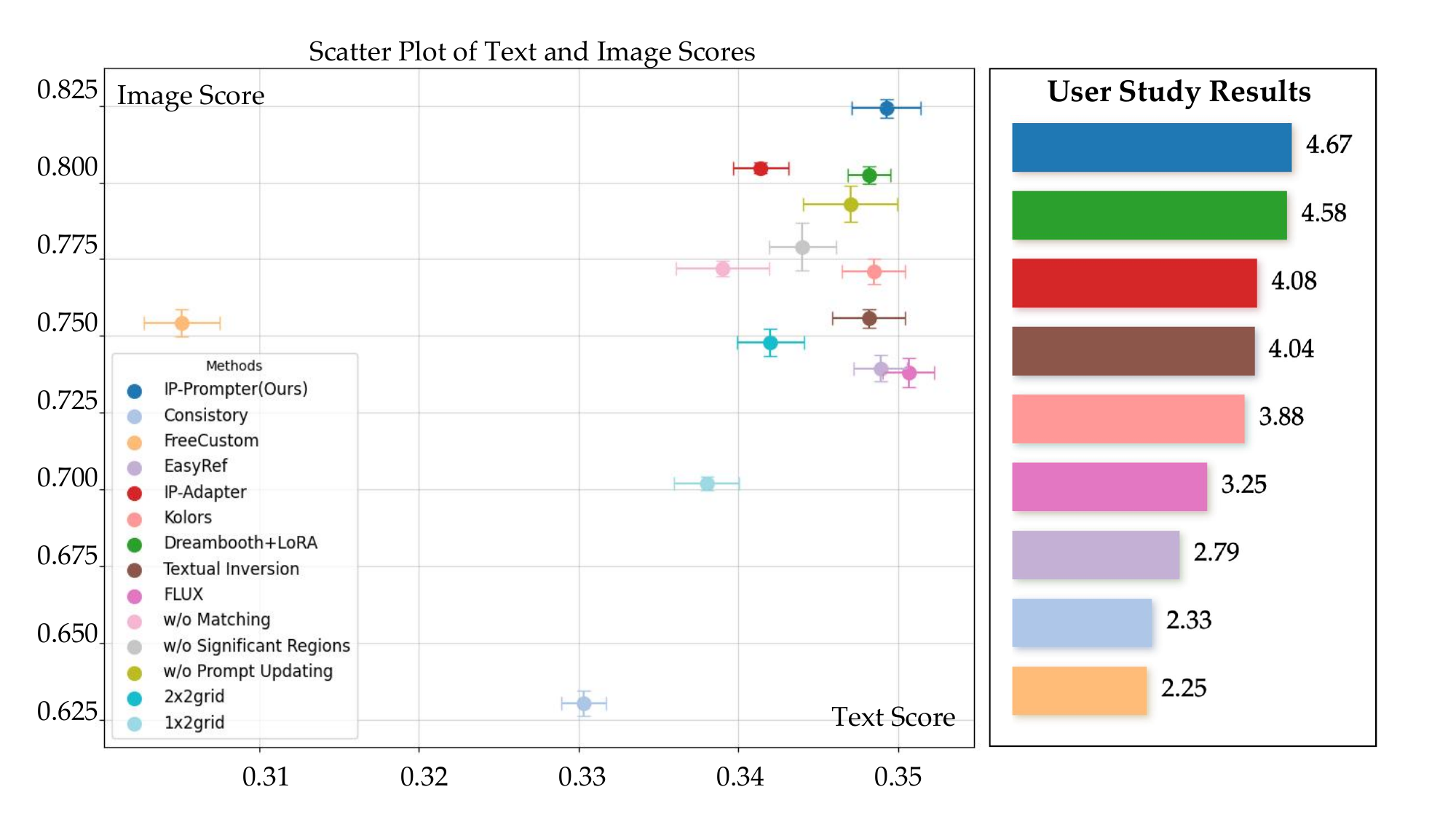}
\caption{
\revision{
\textbf{Quantitative evaluation and user study results.} \sysname achieves comparable scores to the fine-tuned FLUX model.
}
}
\label{fig:quantitative}
\end{figure}

%% file: Figs/fig_ablation.tex
\begin{figure}
\centering
\includegraphics[width=1\linewidth]{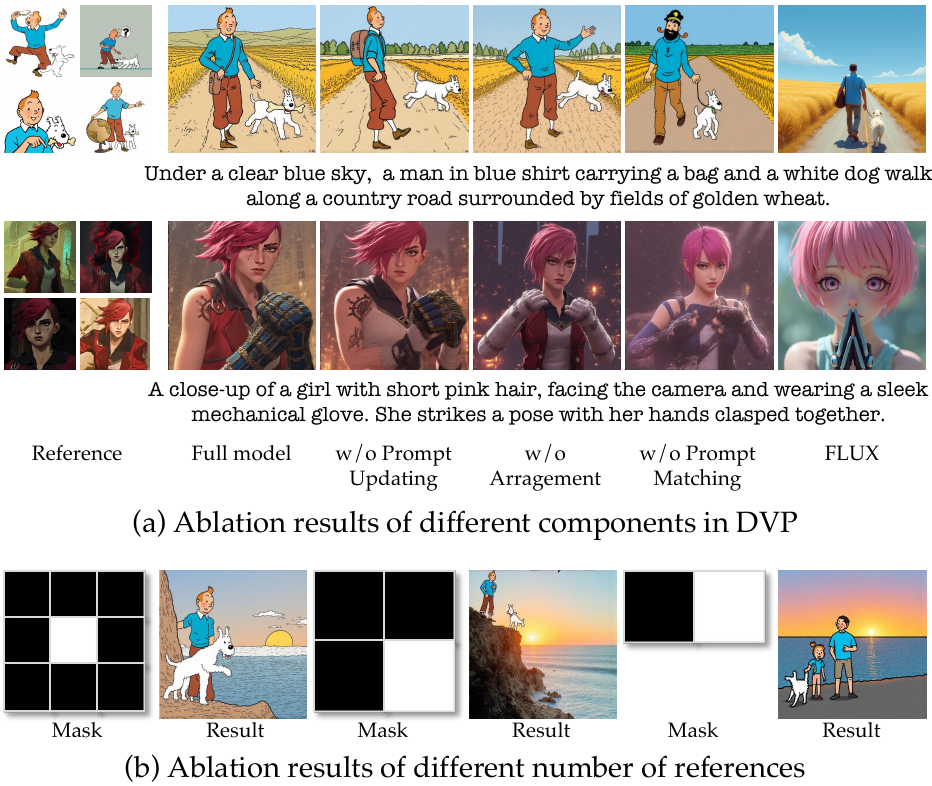}
\caption{Ablation study results.
}
\label{fig:ablation}
\end{figure}

%% file: Figs/fig_app.tex
\begin{figure}
\centering
\includegraphics[width=\linewidth]{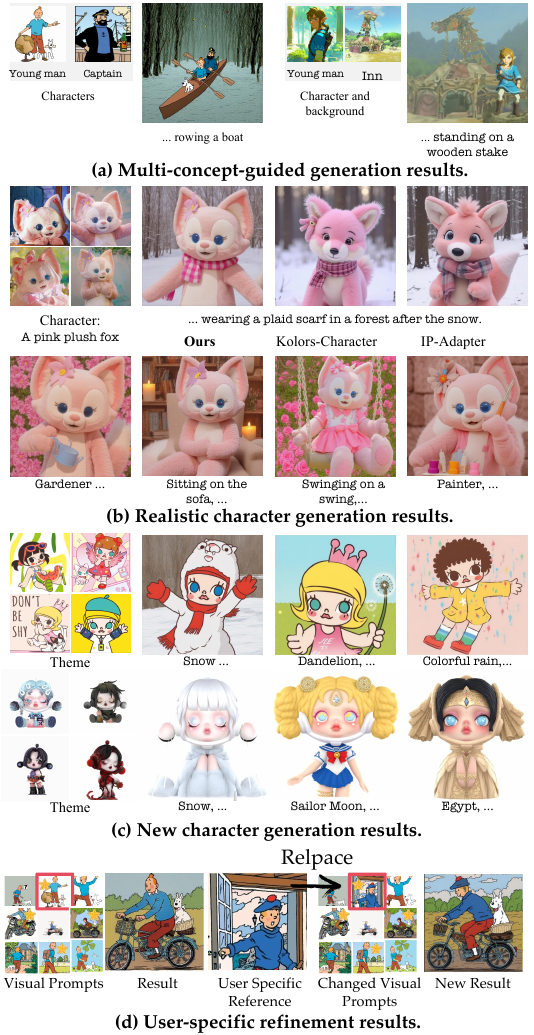}
\caption{Our results of diverse applications.
}
\label{fig:app}
\end{figure}

%% file: Figs/fig_story.tex
\begin{figure*}
\centering
\includegraphics[width=\linewidth]{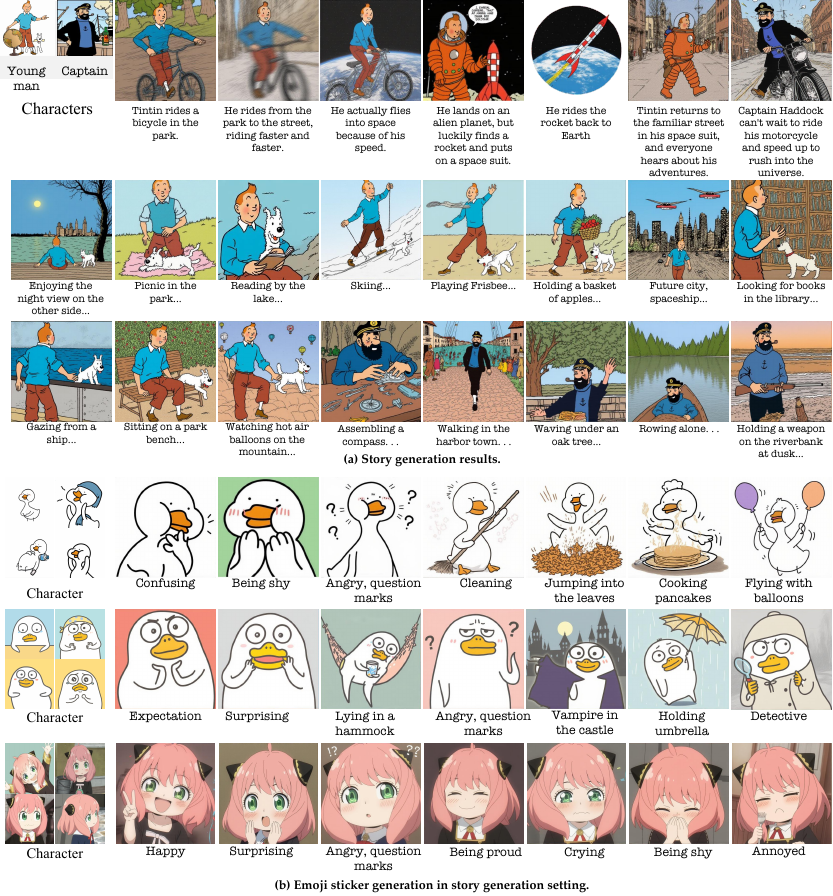}
\caption{Generation results for comic stories and emoji stickers.
}
\label{fig:story}
\end{figure*}

%% file: Figs/fig_story2.tex
\begin{figure*}
\centering
\includegraphics[width=\linewidth]{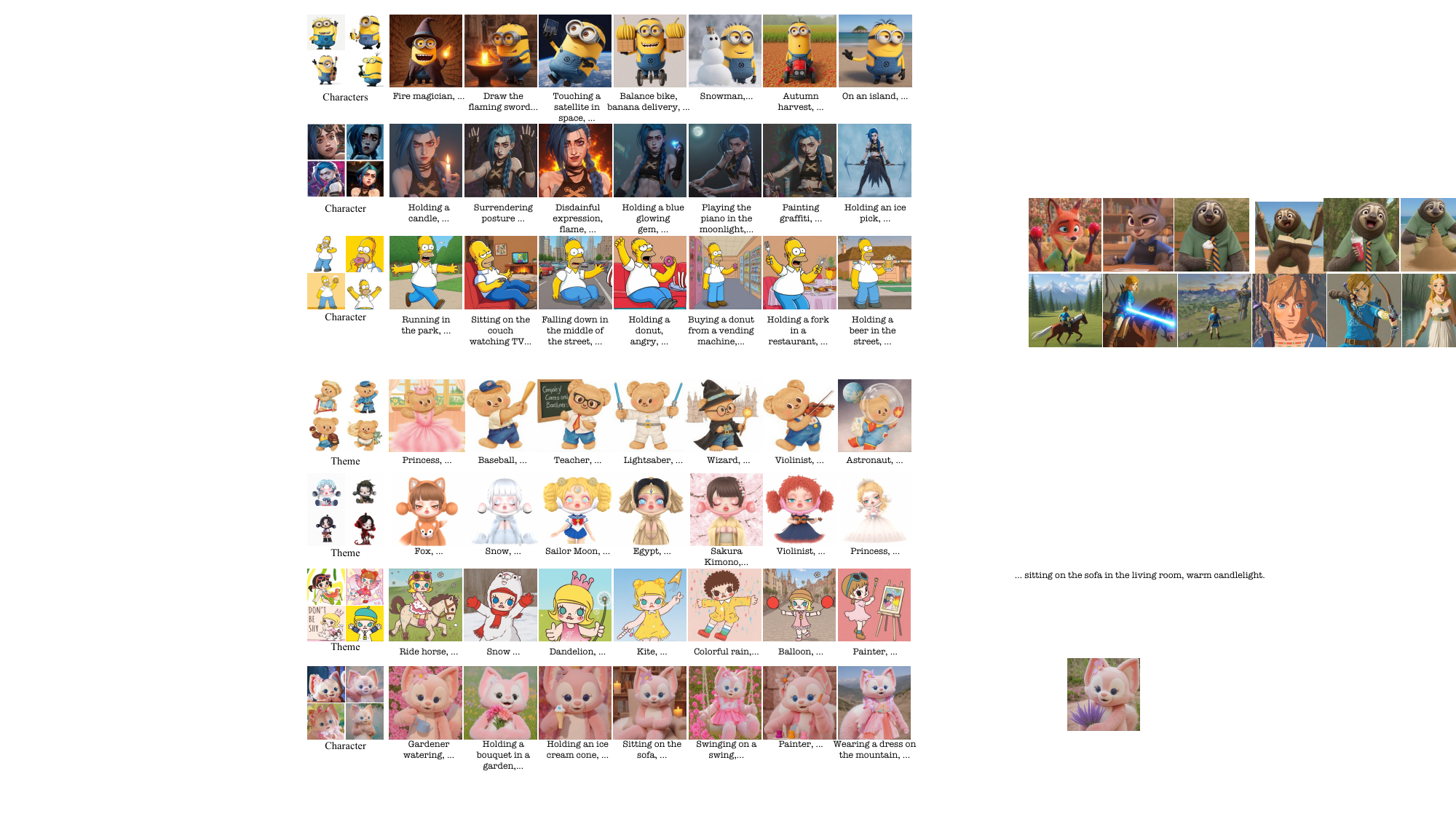}
\caption{More results of story generation.
}
\label{fig:story2}
\end{figure*}

%% file: Sections/5-Application.tex
\section{Applications}

\paragraph{Multi-Concept Generation within a Single Theme}
One core challenge of multi-concept generation is synthesis of images that feature multiple elements, including character interactions, backgrounds, and objects, while maintaining consistency and meaningful interaction between these elements.
Some methods have explored multi-concept generation~\cite{liu2023cones,xu2023versatile,avrahami2023break,customdiff,yeh2024gen4gen,jiang2024mc,gu2024mix,zhang2024spdiffusion,jinimage,kong2024omg}, but these methods require additional training or modification to the network structure.
Fig.~\ref{fig:app}(a) shows the ability of \sysname to generate complex combinations of characters and backgrounds.
\sysname cohesively integrates multiple concepts and facilitates interactions between them, unlocking possibilities for intricate storytelling and character-driven narratives.

\input{Figs/fig_style}

\input{Figs/fig_limitation}
\paragraph{Realistic Character Generation in Photographic Contexts}
\sysname enables the generation of realistic characters seamlessly integrated into photographic backgrounds.
The primary challenge of this task lies in achieving consistent character identity while maintaining editability.
Comparisons with state-of-the-arts methods highlight \sysname’s superior performance in maintaining consistency, as demonstrated in Fig.~\ref{fig:app}(b). These advancements hold considerable potential for applications such as advertisement generation.

\paragraph{New Character Design}
The generation of new characters is less explored~\cite{richardson2024conceptlab}.
\sysname facilitates character design by generating images that adhere to the stylistic and character-specific features of a target theme, while introducing novel concepts.
\sysname's visual prompting template design and DVP facilitates diverse content generation.
As shown in Fig.~\ref{fig:app}(c), experiments validate \sysnames ability to maintain the style and character appearances. 
\sysname provides both professional designers and hobbyists a versatile tool for sparking creativity.

\paragraph{User-Specific Refinement}

As shown in Fig.~\ref{fig:app}(d), DVP allows users to inject personalized images at specified locations in the visual prompts, achieving more refined concept customization.

\paragraph{Consistent Story Generation}
Here we aim to create sequences of images that consistently maintain certain character identities, depict diverse narratives, and facilitate transitions between multiple characters. These are also challenges in the story generation task.
Some methods have explored story generation using text-to-image models~\cite{maharana2022storydall,gong2023talecrafter,he2023animate,ahnKSKLKC23,AutoStory,mao2024story_adapter,zhou2024storymaker,yang2024seedstory,tao2024storyimager,Liu_2024_CVPR,zhou2024storydiffusion}, but these methods require additional training to maintain character identities.
In contrast, \sysname leverages visual prompting to effectively maintain identity within the image domain.
The DVP mechanism enables adaptive matching of visual prompts to accommodate varying plots and characters.
As shown in Figs.~\ref{fig:story} and ~\ref{fig:story2}, \sysname not only creates novel scenes beyond the scope of the original theme, such as characters ``riding in space'', but also seamlessly incorporates iconic visual elements from the source theme, like the ``orange astronaut suit''.
These capabilities highlight the potentiality of \sysname in picture book creation, emoji sticker creation and educational storytelling.

\paragraph{Consistent Style Image Generation}
Artistic style presents distinct challenges in image generation, particularly in maintaining stylistic coherence while introducing diverse content.
Some methods have explored style-guided generation using text-to-image models~\cite{zhang2023inversion,sohn2024styledrop,wang2024instantstyle,chung2024style,hertz2024style,li2024styletokenizer,gao2024styleshot,jeong2024visual}, but these methods often require additional training or modification of the network structure.
As demonstrated in Fig.~\ref{fig:style}, experiments conducted on two distinct artistic styles validate \sysname’s ability to achieve consistent style generation.

%% file: Figs/fig_style.tex
\begin{figure*}
\centering
\includegraphics[width=\linewidth]{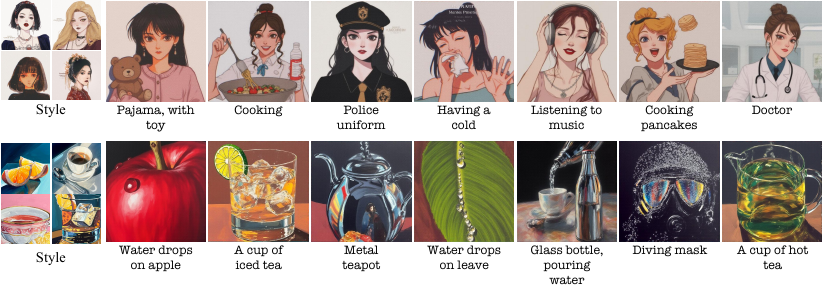}
\caption{Style-guided generation results.
}
\label{fig:style}
\end{figure*}

%% file: Figs/fig_limitation.tex
\begin{figure}
\centering
\includegraphics[width=0.99\linewidth]{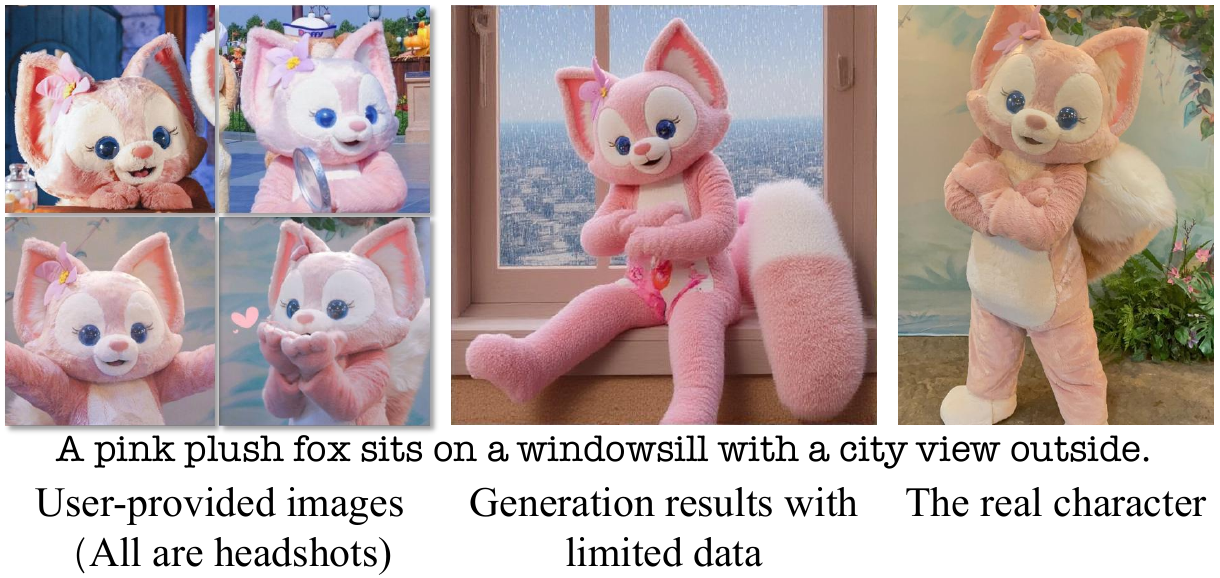}
\caption{Failure cases due to data limitation.
}
\label{fig:limitation}
\end{figure}

%% file: Sections/6-Conclusion.tex
\section{Limitations}
\sysname can generate diverse actions and backgrounds that do not exist in the dataset, but it may be limited by overly homogeneous or insufficient data.
We recommend that users provide images of the target characters in multiple poses and varied backgrounds.
As shown in Fig.~\ref{fig:limitation}, if only facial images of a character are provided and the model is tasked with generating a full-body image, the results may appear reasonable but lack consistency with the actual target.
Beyond collecting richer data, this limitation could be addressed through data augmentation leveraging generative models.

\section{Conclusion}

This paper proposes \sysname, a novel training-free, modification-free method of theme-specific image generation with high flexibility and accuracy. 
We introduce visual prompting, a form of interaction with generative models, which can provide more accurate and direct guidance for models in the visual domain. 
Our dynamic visual prompting pipeline leverages the capabilities of multi-modal models and LLMs while utilizing data-driven advantages to meet the demanding requirements of flexible theme-specific generation.
In comparisons with several state-of-the-art baseline methods, \sysname achieves superior results in both qualitative and quantitative evaluations.
Following a training-free technical paradigm, \sysname delivers performance comparable to fine-tuned models, with the added benefit of producing highly realistic outputs.
Experimental results demonstrate the feasibility and effectiveness of \sysname across a variety of applications, and its low cost lowers the barrier to use.
Additionally, the DVP working mode highlights the potential of inpainting-based visual models in enhancing image generation processes.
We believe that more targeted training and fine-tuning approaches will further unlock the potential of more efficient, controllable generation in future applications.